\documentclass{article}

\PassOptionsToPackage{numbers}{natbib}
 \usepackage[dblblindworkshop, final]{arxiv}
 \usepackage[percent]{overpic}
\workshoptitle{Machine Learning and the Physical Sciences}

\usepackage[utf8]{inputenc} 
\usepackage[T1]{fontenc}    
\usepackage{hyperref}       
\usepackage{url}            
\usepackage{booktabs}       
\usepackage{amsfonts}       
\usepackage{nicefrac}       
\usepackage{microtype}      
\usepackage{xcolor}         
\usepackage{subcaption}
\usepackage{graphicx}
\usepackage{tabularx}
\usepackage{siunitx}

\title{Fine-Tuning Vision-Language Models for Neutrino Event Analysis in High-Energy Physics Experiments}

\author{%
  Dikshant Sagar\thanks{Equal Contribution} \\
  Department of Computer Science\\
  University of California, Irvine\\
  Irvine, CA 92697 \\
  \And
  Kaiwen Yu$^*$ \\
 Department of Computer Science\\
  University of California, Irvine\\
  Irvine, CA 92697 \\
  \And
  Alejandro Yankelevich \\
  Department of Physics \\
  University of California, Irvine\\
  Irvine, CA 92697 \\
  \And
  Jianming Bian \\
  Department of Physics \\
  University of California, Irvine\\
  Irvine, CA 92697 \\
  \And
  Pierre Baldi \\
  Department of Computer Science\\
  University of California, Irvine\\
  Irvine, CA 92697 \\
}

\setcitestyle{square}

\begin{document}

\maketitle

\begin{abstract}

Recent progress in large language models (LLMs)\cite{chang2024survey} has shown strong potential for multimodal reasoning beyond natural language \cite{wu2023multimodal}. In this work, we explore the use of a fine-tuned Vision-Language Model (VLM), based on LLaMA 3.2\cite{grattafiori2024llama}, for classifying neutrino interactions from pixelated detector images in high-energy physics (HEP) experiments. We benchmark its performance against an established CNN baseline used in experiments like NOvA and DUNE\cite{ayres2007nova, falcone2022deep}, evaluating metrics such as classification accuracy, precision, recall, and AUC-ROC. Our results show that the VLM not only matches or exceeds CNN performance but also enables richer reasoning and better integration of auxiliary textual or semantic context. These findings suggest that VLMs offer a promising general-purpose backbone for event classification in HEP, paving the way for multimodal approaches in experimental neutrino physics.
\end{abstract}

\section{Introduction}

Machine learning has become integral to the physical sciences, especially in high-energy physics (HEP), where detectors produce vast, complex datasets \cite{robles2025particle, yankelevich2024sparse, fenton2024reconstruction,baldi2014searching,baldi2016jet,baldi2016parameterized}. Deep learning models are increasingly used to extract meaningful patterns, but their black-box nature often limits interpretability, an essential aspect in scientific analysis\cite{baldi2021deep}. A common task, such as classifying neutrino events against background noise, has traditionally relied on reconstructed detector objects and hand-engineered features, which, while effective, can be limited by reconstruction errors and constrained representations \cite{backhouse2015library}.

This challenge parallels developments in computer vision, where the shift from handcrafted features to deep convolutional neural networks (CNNs) enabled direct learning from raw images and boosted performance \cite{lecun1998convolutional,baldi2021deep}. Inspired by this success, HEP has begun adopting similar approaches, moving toward architectures that can process low-level detector data directly \cite{aurisano2016convolutional,baldi2014searching,baldi2016jet,baldi2016parameterized}. The emergence of Vision Language Models (VLMs), which are large networks pretrained on paired images and text, extends this trend further. VLMs can learn joint visual-semantic representations, offering improved classification capabilities along with interpretable, language-based explanations for predictions \cite{zhang2024vision}.

In this work, we explore fine-tuning VLMs for neutrino event classification. By leveraging their ability to extract features from raw detector inputs, we reduce dependence on engineered variables and demonstrate enhanced classification accuracy over conventional CNNs. Moreover, VLMs provide a richer interpretive layer by generating natural-language rationales, offering a promising path toward more transparent and informative machine learning models in experimental physics. In particular, we compare their performance against conventional CNNs and demonstrate that VLMs not only achieve superior classification accuracy but also provide a broader scope of reasoning and more informative explanations for their predictions.

\section{Methods}

\subsection{Dataset}

We simulate a modular liquid argon time projection chamber (LArTPC) with \SI{5}{mm} pixel readout in a \SI{2}{m}$\times$\SI{2}{m}$\times$\SI{7}{m} volume, featuring alternating anodes and cathodes along the $x$-axis with \SI{0.3}{m} drift lengths. Using GENIE (v3.0.6)~\cite{Andreopoulos:2009rq,Andreopoulos:2015wxa}, 190,000  electron neutrino ($\nu_e$) and muon neutrino ($\nu_\mu$) events with energies up to \SI{10}{GeV} are generated with 74\% interacting through the charged current (CC) and 26\% through the neutral current (NC). Energy deposition is then simulated with GEANT4 (v11.2.0)\cite{Geant:2017ats, Agostinelli:2002hh}. XZ and YZ views are then downsampled to $\SI{5}{cm} \times \SI{5}{cm}$ pixels. Finally, we crop each event display to a $512 \times 512$ grayscale image (``pixel map'') centered on the interaction, creating the final dataset for training.

\subsection{LLaMa 3.2 Vision}
\begin{figure}[!h]
    \centering
    \includegraphics[width=1.\linewidth]{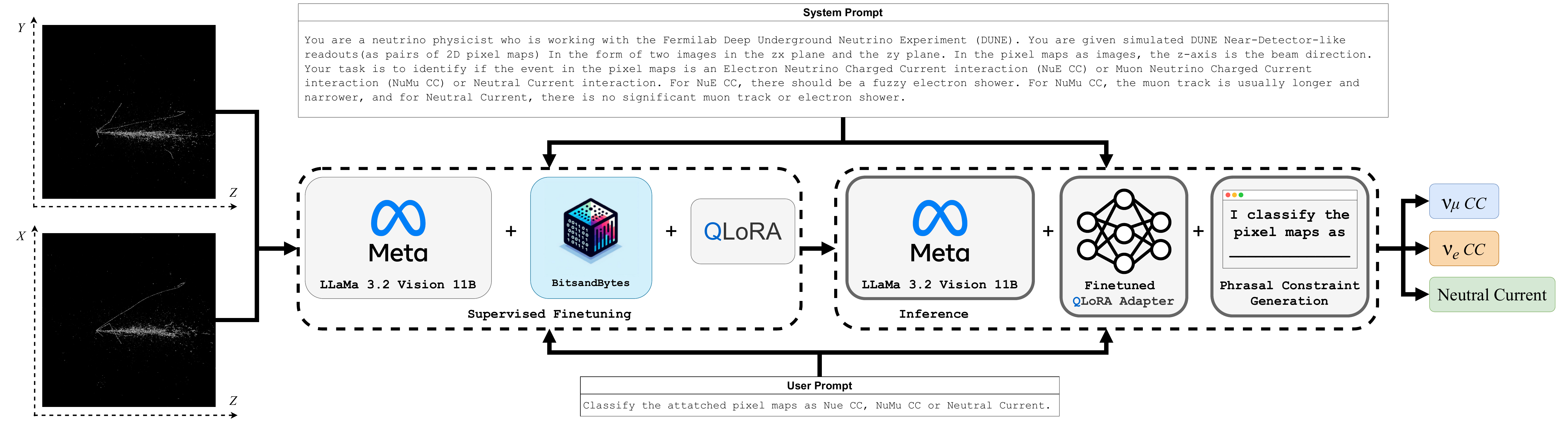}
    \caption{LLaMa 3.2 Vision 11B finetuning pipeline.}
    \label{fig:llamapipeline}
\end{figure}

LLaMA Vision 3.2 is a family of VLMs developed by Meta~\cite{grattafiori2024llama}, extending the LLaMA 3.2 series with visual understanding capabilities. Unlike traditional CNNs designed solely for image-based tasks, LLaMA Vision integrates visual and textual modalities within a unified transformer architecture~\cite{vaswani2017attention}. Trained on a broad corpus of images and documents, it can process inputs ranging from photographs and plots to pixelated detector data alongside natural language.

The model employs a high-resolution vision encoder that tokenizes images into patch embeddings~\cite{dosovitskiy2020image}, which are jointly processed with text tokens by a shared transformer decoder. This enables cross-modal contextual reasoning, making it particularly suited for tasks like neutrino event classification using sparse detector images. In this work, we fine-tune the 11B parameter version using supervised instruction tuning and QLoRA~\cite{dettmers2023qlora}, as outlined in Fig.~\ref{fig:llamapipeline}. This enables the model to adapt to physics-specific data while preserving its pretrained multimodal capabilities.

A key advantage of this setup is the model’s ability to generate both predictions and natural-language justifications, improving interpretability and scientific insight. Through this study, we assess whether VLMs like LLaMA Vision 3.2 can offer competitive or complementary alternatives to traditional CNNs in high-energy physics applications.

\subsubsection{Parameter Efficient Supervised Finetuning}

Fine-tuning VLMs like LLaMA Vision 3.2~\cite{grattafiori2024llama} is computationally demanding due to their billions of parameters and memory footprint. Fully updating all model weights is often infeasible for domain-specific datasets, such as those in neutrino physics, which are typically small and risk overfitting or catastrophic forgetting~\cite{hu2022lora}. To address this, we adopt a parameter-efficient fine-tuning (PEFT) strategy, using Quantized Low-Rank Adaptation (QLoRA) ~\cite{dettmers2023qlora}. QLoRA combines two key techniques: (1) quantization, where base model weights are stored in 4-bit precision to reduce memory consumption; and (2) low-rank adaptation~\cite{hu2022lora}, where small trainable matrices are added to attention and MLP layers, enabling effective task-specific learning without updating the full model. This approach allows us to fine-tune LLaMA Vision 3.2 11B~\cite{grattafiori2024llama} on our neutrino dataset using modest GPU resources, while preserving the model's visual-linguistic capabilities. It enables faster experimentation, lower hardware demands, and makes large-model deployment more practical for high-energy physics research.

\subsubsection{Model and Training Specifications}

We fine-tune the LLaMA 3.2 Vision Instruct 11B model. Its instruction-following capabilities allow us to frame neutrino event classification as a multimodal prompt-response task. Using the \texttt{meta-llama/Llama-3.2-11B-Vision-Instruct} checkpoint with 4-bit quantization via the \texttt{BitsAndBytes} library, we significantly reduce memory usage during training. Fine-tuning is conducted on four NVIDIA A6000 GPUs (49GB VRAM each) with a per-device batch size of 4. To enable efficient adaptation, we employ QLoRA~\cite{dettmers2023qlora}, which introduces low-rank trainable matrices into key network modules while keeping the base model frozen. Supervised fine-tuning is performed using Hugging Face's \texttt{SFTTrainer}~\cite{wolf2019huggingface} on 190,000 labeled events over one epoch. The training setup includes an effective batch size of 8, the \texttt{adamw\_torch\_fused} optimizer with a learning rate of $2 \times 10^{-4}$, warm-up ratio of 0.03, and gradient clipping at 0.3. We use \texttt{bfloat16} precision with \texttt{TF32} fallback for numerical stability. Training was completed in approximately one week, with regular checkpointing and logging via TensorBoard.

\subsubsection{Inference}

During inference, the model was prompted with a standardized system message outlining physics-specific context and interaction class descriptions, along with a user instruction to classify each event as $\nu_{e}$ CC, $\nu_{\mu}$ CC, or NC, mirroring the fine-tuning setup (Fig.~\ref{fig:llamapipeline}). VLMs generate text autoregressively, predicting one token at a time based on the prompt and previous tokens~\cite{chang2024survey,wu2023multimodal}. However, unconstrained generation can lead to inconsistent phrasing, complicating evaluation. To enforce consistency, we applied phrasal constraints~\cite{hokamp2017lexically}, requiring outputs to begin with “I classify the pixel maps as,” followed by one of the canonical class labels. This was implemented via constrained beam search using token ID prefixes, ensuring machine-readable, deterministic output while retaining the model’s full visual-linguistic reasoning capacity. This approach simplifies confidence extraction, enhances reproducibility, and reduces output variability. To quantify model confidence, we extracted log-softmax scores for the first token of each class label immediately after the fixed prompt prefix. These log-probabilities reflect the model’s relative preference for each class~\cite{petroni2019language}. We then applied a temperature-scaled softmax transformation to compute normalized confidence scores, $P(C_i) = \text{softmax}(T \cdot \log p(C_i)) $ with temperature $T = 5$, where $P(C_i)$ is the probability assigned to class $C_i$. This yields interpretable class probabilities that emphasize the most likely prediction while preserving information about alternative choices~\cite{guo2017calibration}.

\subsection{CNN Baseline}

Convolutional neural networks (CNNs) have been effective in high-energy physics tasks such as event classification and image segmentation\cite{refId0,aurisano2016convolutional, robles2025particle}, thanks to their ability to model local spatial features. Similar to previous works, we implemented a lightweight Siamese-style\cite{bromley1993signature, koch2015siamese} CNN with approximately 3.4M parameters as a baseline for neutrino event classification. Each input image is processed through identical convolutional branches that are later merged for joint inference. This architecture is optimized for high-resolution, sparsely populated pixel maps commonly found in neutrino detectors. Each branch begins with a ReLU6-activated convolution\cite{agarap2018deep}, followed by a series of inverted residual blocks adapted from MobileNetV2\cite{sandler2018mobilenetv2}, with varying expansion rates, kernel sizes, and output channels. Some blocks incorporate Squeeze-and-Excitation (SE) modules \cite{hu2018squeeze} for channel-wise attention, and custom activations (ReLU6, hard-swish) are used throughout. The two feature streams are concatenated and further processed via additional inverted residual blocks\cite{sandler2018mobilenetv2}, followed by global average pooling, dropout\cite{srivastava2014dropout}, and fully connected layers. The model outputs probabilities over three neutrino interaction classes. 


\subsubsection{Training Setup}


We train the CNN baseline in a supervised setting using the same $2 \times 512 \times 512$ grayscale detector images, each representing a neutrino interaction. The Siamese architecture processes each image independently before joint classification. Training uses the Adam optimizer\cite{kingma2014adam} (learning rate $1 \times 10^{-6}$), with a batch size of 16. We train for up to 300 epochs, applying early stopping if validation loss plateaus for 10 epochs. The objective is cross-entropy loss. All CNN experiments are conducted using PyTorch on a single NVIDIA A5000 GPU.

\section{Results}

\begin{figure}[!h]
    \centering
    \begin{subfigure}{0.32\textwidth}
        \centering
        \begin{overpic}[width=\linewidth]
        {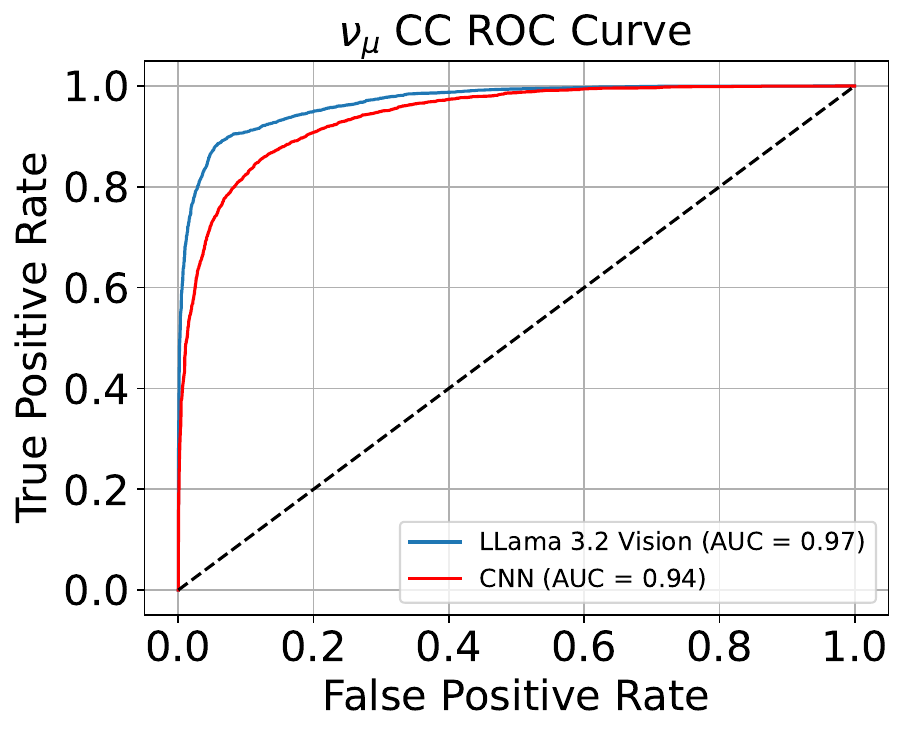}
            \put(90, -2){\small (a)}
        \end{overpic}
    \end{subfigure}
    \begin{subfigure}{0.32\textwidth}
        \centering
        \begin{overpic}[width=\linewidth]
        {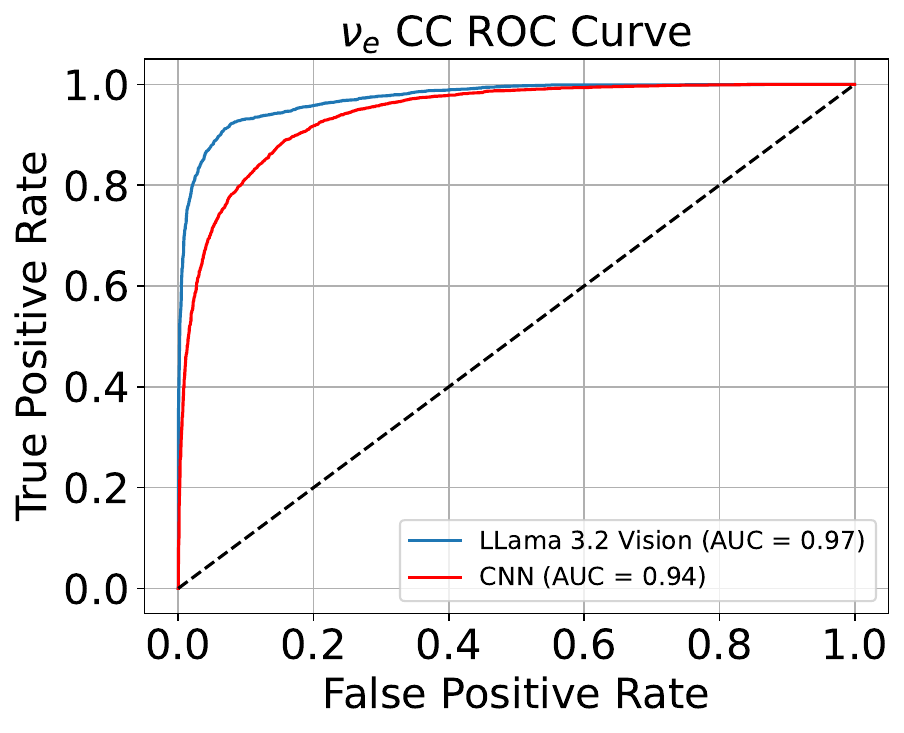}
            \put(90, -2){\small (b)}
        \end{overpic}
    \end{subfigure}
    \begin{subfigure}{0.32\textwidth}
        \centering
        \begin{overpic}[width=\linewidth]
        {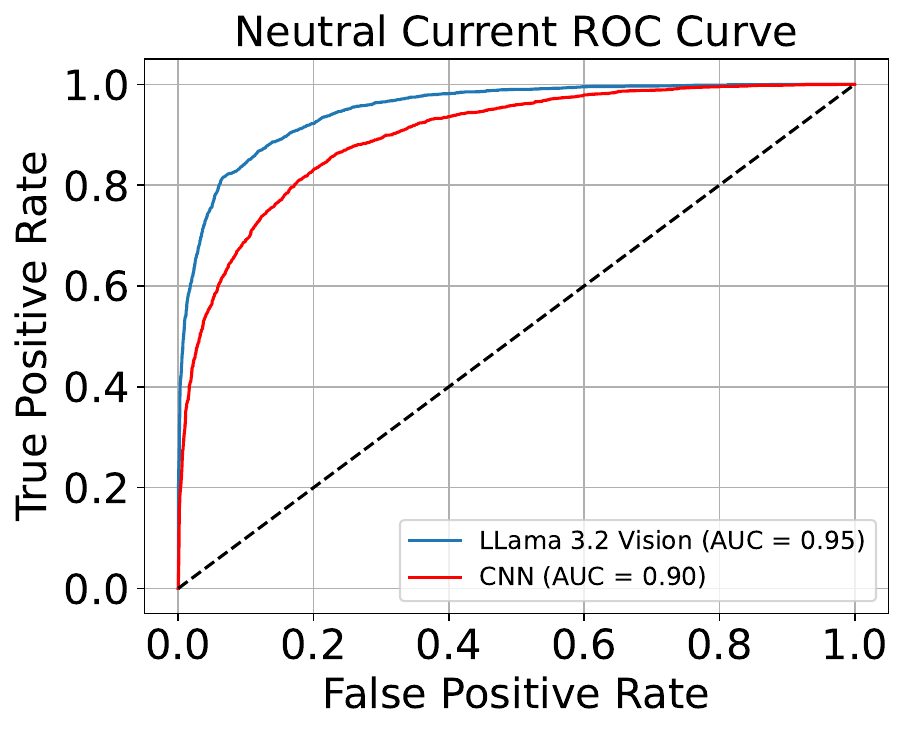}
            \put(90, -2){\small (c)}
        \end{overpic}
    \end{subfigure}
    \caption{AUC-ROC curves for each class (a) $\nu_\mu$ CC, (b) $\nu_e$ CC, and (c) NC comparing performance between the finetuned LLaMa 3.2 Vision and the CNN.}
    \label{fig:roc-curves}
\end{figure}

For evaluation, we performed inference on a held-out test set comprising 5\% of the data (10,000 events). The LLaMA 3.2 Vision-Instruct model demonstrated superior performance in classifying neutrino interaction events from simulated detector pixel maps, consistently outperforming a conventional CNN baseline across key metrics. As shown in Table~\ref{metrics_table}, LLaMA achieved higher accuracy (0.87), precision and recall (0.87), and AUC-ROC (0.96) (See Figure. \ref{fig:roc-curves}), compared to the CNN’s lower accuracy (0.68), precision and recall (0.78), and AUC-ROC (0.93) (See Figure. \ref{fig:roc-curves}).

\begin{figure}[!h]
    \centering
    \begin{subfigure}{0.35\textwidth}
        \centering
        \includegraphics[width=\linewidth]{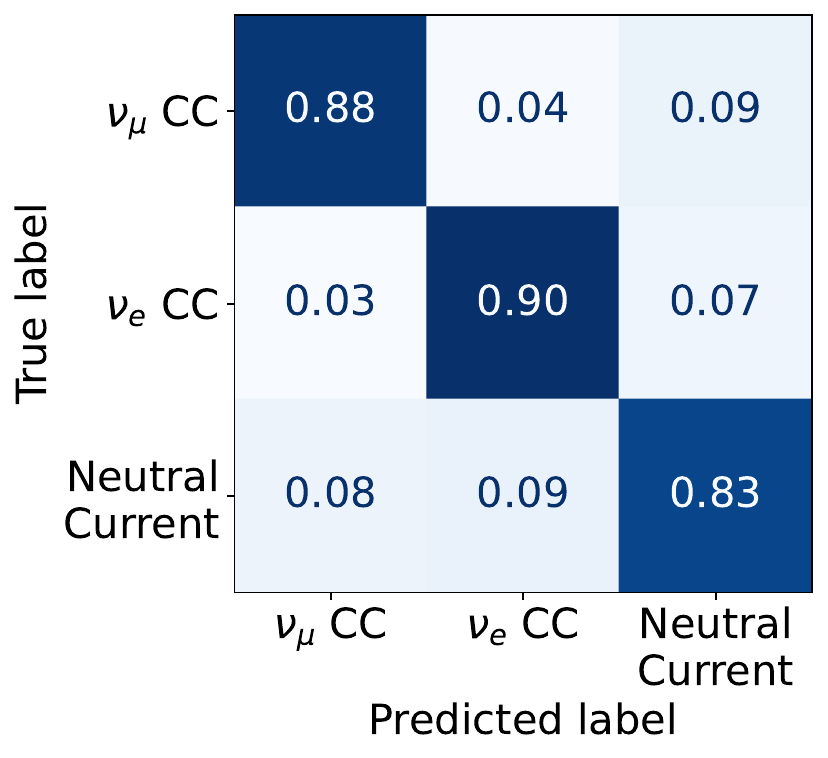}
        \caption{}
    \end{subfigure}
    \begin{subfigure}{0.35\textwidth}
        \centering
        \includegraphics[width=\linewidth]{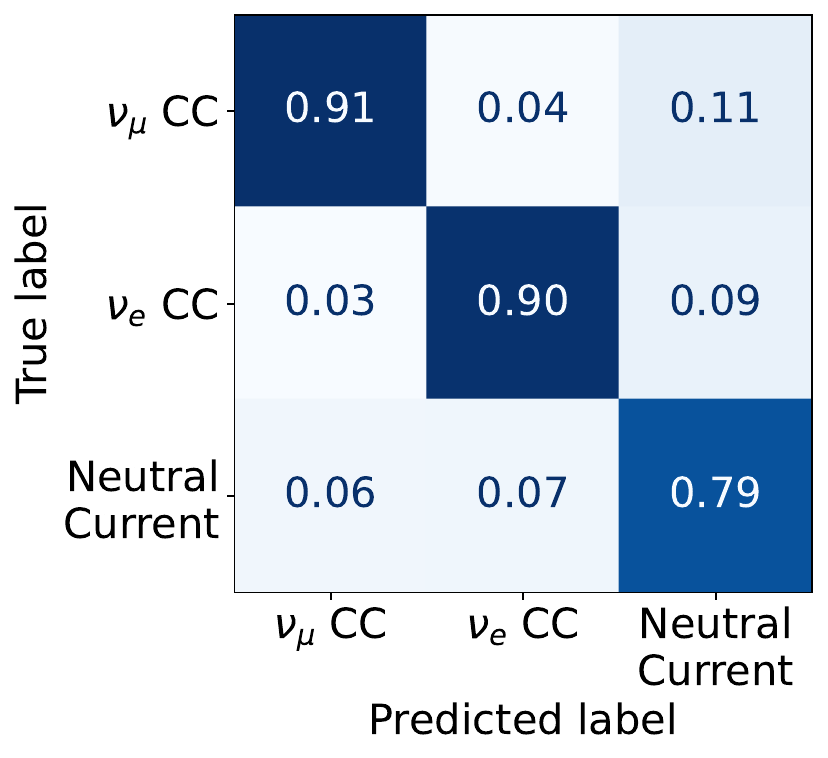}
        \caption{}
    \end{subfigure}\\
    \begin{subfigure}{0.35\textwidth}
        \centering
        \includegraphics[width=\linewidth]{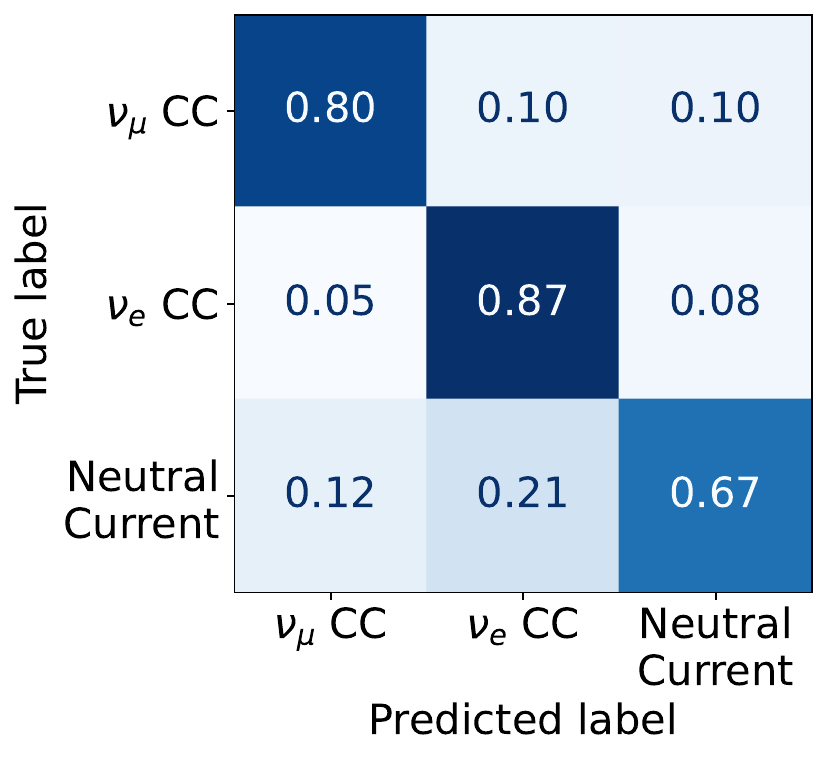}
        \caption{}
    \end{subfigure}
    \begin{subfigure}{0.35\textwidth}
        \centering
        \includegraphics[width=\linewidth]{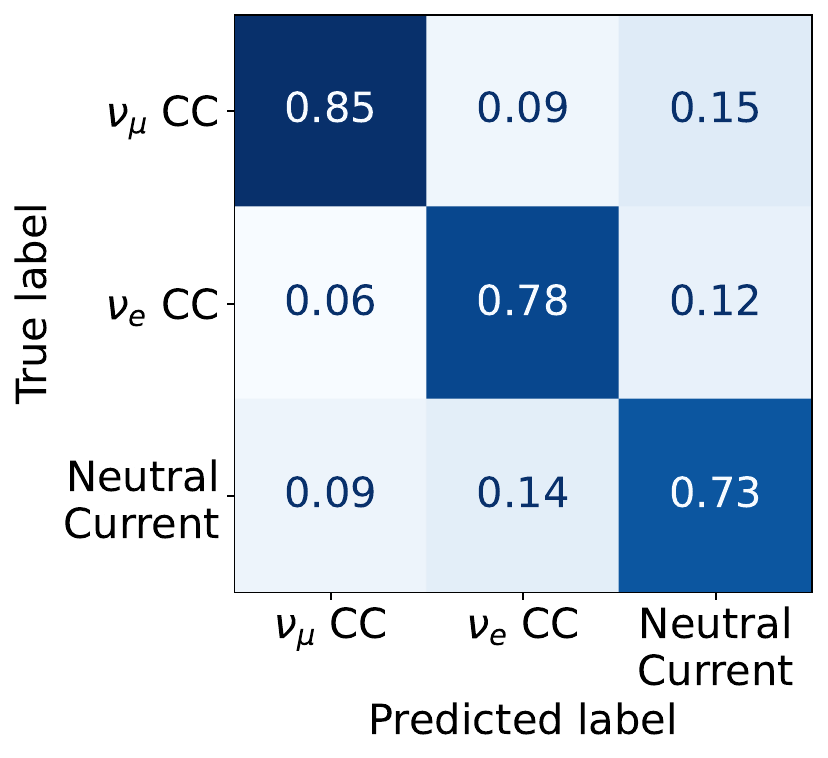}
        \caption{}
    \end{subfigure}
    \caption{Finetuned LLaMa's (a) recall matrix (truth normalized) and (b) precision matrix (prediction normalized) and CNN's (c) recall matrix (truth normalized) and (d) precision matrix (prediction normalized).}
    \label{fig:llama-cm}
\end{figure}

The confusion matrices in Figure~\ref{fig:llama-cm} show that LLaMA also offers greater class-wise precision and recall across all three classes. Its performance in NC identification is especially improved with better $\nu_{e}$ CC vs NC discrimination, which is particularly important for neutrino physics experiments. Its balanced precision and recall also contributes to more reliable classification behavior. The ROC curves in Figure~\ref{fig:roc-curves} show that LLaMA maintains this performance across a wide range of acceptance thresholds.

These gains come with higher resource requirements: LLaMA consumes 25.4~GB of memory per inference and requires 3.3 seconds per sample, compared to 1.0~GB and 20 milliseconds for the CNN. Nevertheless, the improved accuracy and interpretability offered by LLaMA make it a powerful alternative for scientific applications where precision is paramount and computational resources are available.




\begin{table}[!h]
    \caption{Event classification aggregated metrics.}
    \label{metrics_table}
    \centering
    \begin{tabular}{lcc}
    \toprule
        Metric & LLaMa 3.2 Vision & CNN\\
    \midrule
         Accuracy & \textbf{0.87} & 0.68 \\
         Precision & \textbf{0.87} & 0.78 \\
         Recall &\textbf{0.87} & 0.78 \\
         AUC-ROC & \textbf{0.96} & 0.93\\
     \midrule
         Inference Memory Usage (MB) & 25412.91 & \textbf{1000.57}\\
         Time per Sample (mSec) & 3300 & \textbf{20}\\
    \bottomrule
    \end{tabular}
\end{table}


\section{Conclusion}
We compared a lightweight CNN with LLaMA 3.2 Vision for neutrino event classification and found that while LLaMA incurs substantially higher memory and inference costs, it delivers superior accuracy and the added benefit of interpretability through textual explanations. This makes VLMs well-suited for offline scientific analysis, where explainability is crucial. CNNs, by contrast, are more appropriate for real-time or resource-limited settings. Future work includes compressing large models, distilling them into efficient variants, and building domain-specific foundation models tailored to neutrino physics.

\bibliography{references}

\begin{thebibliography}{10}

\bibitem{agarap2018deep}
{\sc Agarap, A.~F.}
\newblock Deep learning using rectified linear units (relu).
\newblock {\em arXiv preprint arXiv:1803.08375\/} (2018).

\bibitem{Agostinelli:2002hh}
{\sc Agostinelli, S., et~al.}
\newblock {GEANT4--a simulation toolkit}.
\newblock {\em Nucl. Instrum. Meth. A 506\/} (2003), 250--303.

\bibitem{Andreopoulos:2015wxa}
{\sc Andreopoulos, C., Barry, C., Dytman, S., Gallagher, H., Golan, T., Hatcher, R., Perdue, G., and Yarba, J.}
\newblock {The GENIE Neutrino Monte Carlo Generator: Physics and User Manual}, Oct 2015.

\bibitem{Andreopoulos:2009rq}
{\sc Andreopoulos, C., et~al.}
\newblock {The GENIE Neutrino Monte Carlo Generator}.
\newblock {\em Nucl. Instrum. Meth. A 614\/} (2010), 87--104.

\bibitem{aurisano2016convolutional}
{\sc Aurisano, A., Radovic, A., Rocco, D., Himmel, A., Messier, M., Niner, E., Pawloski, G., Psihas, F., Sousa, A., and Vahle, P.}
\newblock A convolutional neural network neutrino event classifier.
\newblock {\em Journal of Instrumentation 11}, 09 (2016), P09001.

\bibitem{ayres2007nova}
{\sc Ayres, D., Drake, G., Goodman, M., Grudzinski, J., Guarino, V., Talaga, R., Zhao, A., Stamoulis, P., Stiliaris, E., Tzanakos, G., et~al.}
\newblock The nova technical design report.

\bibitem{backhouse2015library}
{\sc Backhouse, C., and Patterson, R.}
\newblock Library event matching event classification algorithm for electron neutrino interactions in the no$\nu$a detectors.
\newblock {\em Nuclear Instruments and Methods in Physics Research Section A: Accelerators, Spectrometers, Detectors and Associated Equipment 778\/} (2015), 31--39.

\bibitem{baldi2021deep}
{\sc Baldi, P.}
\newblock {\em Deep learning in science}.
\newblock Cambridge University Press, 2021.

\bibitem{baldi2016jet}
{\sc Baldi, P., Bauer, K., Eng, C., Sadowski, P., and Whiteson, D.}
\newblock Jet substructure classification in high-energy physics with deep neural networks.
\newblock {\em Physical Review D 93}, 9 (2016), 094034.

\bibitem{baldi2016parameterized}
{\sc Baldi, P., Cranmer, K., Faucett, T., Sadowski, P., and Whiteson, D.}
\newblock Parameterized neural networks for high-energy physics.
\newblock {\em The European Physical Journal C 76}, 5 (2016), 1--7.

\bibitem{baldi2014searching}
{\sc Baldi, P., Sadowski, P., and Whiteson, D.}
\newblock Searching for exotic particles in high-energy physics with deep learning.
\newblock {\em Nature communications 5}, 1 (2014), 4308.

\bibitem{bromley1993signature}
{\sc Bromley, J., Guyon, I., LeCun, Y., S{\"a}ckinger, E., and Shah, R.}
\newblock Signature verification using a" siamese" time delay neural network.
\newblock {\em Advances in neural information processing systems 6\/} (1993).

\bibitem{chang2024survey}
{\sc Chang, Y., Wang, X., Wang, J., Wu, Y., Yang, L., Zhu, K., Chen, H., Yi, X., Wang, C., Wang, Y., et~al.}
\newblock A survey on evaluation of large language models.
\newblock {\em ACM transactions on intelligent systems and technology 15}, 3 (2024), 1--45.

\bibitem{dettmers2023qlora}
{\sc Dettmers, T., Pagnoni, A., Holtzman, A., and Zettlemoyer, L.}
\newblock Qlora: Efficient finetuning of quantized llms.
\newblock {\em Advances in neural information processing systems 36\/} (2023), 10088--10115.

\bibitem{dosovitskiy2020image}
{\sc Dosovitskiy, A., Beyer, L., Kolesnikov, A., Weissenborn, D., Zhai, X., Unterthiner, T., Dehghani, M., Minderer, M., Heigold, G., Gelly, S., et~al.}
\newblock An image is worth 16x16 words: Transformers for image recognition at scale.
\newblock {\em arXiv preprint arXiv:2010.11929\/} (2020).

\bibitem{falcone2022deep}
{\sc Falcone, A., Collaboration, D., et~al.}
\newblock Deep underground neutrino experiment: Dune.
\newblock {\em Nuclear Instruments and Methods in Physics Research Section A: Accelerators, Spectrometers, Detectors and Associated Equipment 1041\/} (2022), 167217.

\bibitem{fenton2024reconstruction}
{\sc Fenton, M.~J., Shmakov, A., Okawa, H., Li, Y., Hsiao, K.-Y., Hsu, S.-C., Whiteson, D., and Baldi, P.}
\newblock Reconstruction of unstable heavy particles using deep symmetry-preserving attention networks.
\newblock {\em Communications Physics 7}, 1 (2024), 139.

\bibitem{Geant:2017ats}
{\sc {Geant4 Collaboration}}.
\newblock Geant4 10.4 release notes.
\newblock {\em geant4-data.web.cern.ch\/} (2017).

\bibitem{grattafiori2024llama}
{\sc Grattafiori, A., Dubey, A., Jauhri, A., Pandey, A., Kadian, A., Al-Dahle, A., Letman, A., Mathur, A., Schelten, A., Vaughan, A., et~al.}
\newblock The llama 3 herd of models.
\newblock {\em arXiv preprint arXiv:2407.21783\/} (2024).

\bibitem{guo2017calibration}
{\sc Guo, C., Pleiss, G., Sun, Y., and Weinberger, K.~Q.}
\newblock On calibration of modern neural networks.
\newblock In {\em International conference on machine learning\/} (2017), PMLR, pp.~1321--1330.

\bibitem{hokamp2017lexically}
{\sc Hokamp, C., and Liu, Q.}
\newblock Lexically constrained decoding for sequence generation using grid beam search.
\newblock {\em arXiv preprint arXiv:1704.07138\/} (2017).

\bibitem{hu2022lora}
{\sc Hu, E.~J., Shen, Y., Wallis, P., Allen-Zhu, Z., Li, Y., Wang, S., Wang, L., Chen, W., et~al.}
\newblock Lora: Low-rank adaptation of large language models.
\newblock {\em ICLR 1}, 2 (2022), 3.

\bibitem{hu2018squeeze}
{\sc Hu, J., Shen, L., and Sun, G.}
\newblock Squeeze-and-excitation networks.
\newblock In {\em Proceedings of the IEEE conference on computer vision and pattern recognition\/} (2018), pp.~7132--7141.

\bibitem{kingma2014adam}
{\sc Kingma, D.~P., and Ba, J.}
\newblock Adam: A method for stochastic optimization.
\newblock {\em arXiv preprint arXiv:1412.6980\/} (2014).

\bibitem{koch2015siamese}
{\sc Koch, G., Zemel, R., Salakhutdinov, R., et~al.}
\newblock Siamese neural networks for one-shot image recognition.
\newblock In {\em ICML deep learning workshop\/} (2015), vol.~2, Lille, pp.~1--30.

\bibitem{lecun1998convolutional}
{\sc LeCun, Y., and Bengio, Y.}
\newblock Convolutional networks for images, speech, and time series.
\newblock {\em The handbook of brain theory and neural networks\/} (1998).

\bibitem{refId0}
{\sc {Madrazo, Celia Fern{\'a}ndez}, {Heredia, Ignacio}, {Lloret, Lara}, and {Marco de Lucas, Jes{\'u}s}}.
\newblock Application of a convolutional neural network for image classification for the analysis of collisions in high energy physics.
\newblock {\em EPJ Web Conf. 214\/} (2019), 06017.

\bibitem{petroni2019language}
{\sc Petroni, F., Rockt{\"a}schel, T., Lewis, P., Bakhtin, A., Wu, Y., Miller, A.~H., and Riedel, S.}
\newblock Language models as knowledge bases?
\newblock {\em arXiv preprint arXiv:1909.01066\/} (2019).

\bibitem{robles2025particle}
{\sc Robles, E.~E., Yankelevich, A., Wu, W., Bian, J., and Baldi, P.}
\newblock Particle hit clustering and identification using point set transformers in liquid argon time projection chambers.
\newblock {\em Journal of Instrumentation 20}, 07 (2025), P07030.

\bibitem{sandler2018mobilenetv2}
{\sc Sandler, M., Howard, A., Zhu, M., Zhmoginov, A., and Chen, L.-C.}
\newblock Mobilenetv2: Inverted residuals and linear bottlenecks.
\newblock In {\em Proceedings of the IEEE conference on computer vision and pattern recognition\/} (2018), pp.~4510--4520.

\bibitem{srivastava2014dropout}
{\sc Srivastava, N., Hinton, G., Krizhevsky, A., Sutskever, I., and Salakhutdinov, R.}
\newblock Dropout: a simple way to prevent neural networks from overfitting.
\newblock {\em The journal of machine learning research 15}, 1 (2014), 1929--1958.

\bibitem{vaswani2017attention}
{\sc Vaswani, A., Shazeer, N., Parmar, N., Uszkoreit, J., Jones, L., Gomez, A.~N., Kaiser, {\L}., and Polosukhin, I.}
\newblock Attention is all you need.
\newblock {\em Advances in neural information processing systems 30\/} (2017).

\bibitem{wolf2019huggingface}
{\sc Wolf, T., Debut, L., Sanh, V., Chaumond, J., Delangue, C., Moi, A., Cistac, P., Rault, T., Louf, R., Funtowicz, M., et~al.}
\newblock Huggingface's transformers: State-of-the-art natural language processing.
\newblock {\em arXiv preprint arXiv:1910.03771\/} (2019).

\bibitem{wu2023multimodal}
{\sc Wu, J., Gan, W., Chen, Z., Wan, S., and Yu, P.~S.}
\newblock Multimodal large language models: A survey.
\newblock In {\em 2023 IEEE International Conference on Big Data (BigData)\/} (2023), IEEE, pp.~2247--2256.

\bibitem{yankelevich2024sparse}
{\sc Yankelevich, A., Shmakov, A., Bian, J., and Baldi, P.}
\newblock Sparse convolution transformers for dune fd event and particle classification.
\newblock {\em Bulletin of the American Physical Society\/} (2024).

\bibitem{zhang2024vision}
{\sc Zhang, J., Huang, J., Jin, S., and Lu, S.}
\newblock Vision-language models for vision tasks: A survey.
\newblock {\em IEEE transactions on pattern analysis and machine intelligence 46}, 8 (2024), 5625--5644.

\end{thebibliography}
\end{document}